\documentclass[copyright,creativecommons]{eptcs}
\providecommand{\event}{DCM 2010} 
\usepackage{breakurl}             
\usepackage[latin1]{inputenc}
\usepackage{graphicx}
\usepackage{algorithmic}
\usepackage{algorithm}
\usepackage{url}

\title{Towards the Design of Heuristics by Means of Self-Assembly}
\author{German Terrazas
\institute{ASAP Group, School of Computer Science\\
University of Nottingham, UK}
\email{gzt@cs.nott.ac.uk}
\and
Dario Landa-Silva
\institute{ASAP Group, School of Computer Science\\
University of Nottingham, UK}
\email{jds@cs.nott.ac.uk}
\and
 Natalio Krasnogor
\institute{ASAP Group, School of Computer Science\\
University of Nottingham, UK}
\email{nxk@cs.nott.ac.uk}
}
\def\titlerunning{Towards the Design of Heuristics by Means of Self-Assembly}
\def\authorrunning{G. Terrazas, D. Landa-Silva \& N. Krasnogor}
\begin{document}
\maketitle

\begin{abstract}
The current investigations on hyper-heuristics design have sprung up in two different flavours: heuristics that choose heuristics and heuristics that generate heuristics. In the latter, the goal is to develop a problem-domain independent strategy to automatically generate a good performing heuristic for the problem at hand. This can be done, for example, by automatically selecting and combining different low-level heuristics into a problem specific and effective strategy. Hyper-heuristics raise the level of generality on automated problem solving by attempting to select and/or generate tailored heuristics for the problem at hand. Some approaches like genetic programming have been proposed for this. In this paper, we explore an elegant nature-inspired alternative based on self-assembly construction processes, in which structures emerge out of local interactions between autonomous components. This idea arises from previous works in which computational models of self-assembly were subject to evolutionary design in order to perform the automatic construction of user-defined structures. Then, the aim of this paper is to present a novel methodology for the automated design of heuristics by means of self-assembly.
\end{abstract}
\section{Introduction}

A \emph{hyper-heuristic} is a search methodology that selects and combines heuristics to generate good solutions for a given problem. The design of a hyper-heuristic is important and we believe that adapting natural construction models is a suitable approach to consider. Cooperative construction processes capable of orchestrating adequate building blocks to achieve efficient composites are observed in nature such as self-assembly and self-organisation. In particular, self-assembly is a phenomenon in which complex structures emerge out of local interactions between autonomous components. The purpose of this paper is to propose a nature-inspired cooperative strategy as a method for the automated construction of heuristic search strategies. Given a computational search problem and a set of simpler heuristics embodied in self-assembly entities, the idea is to develop a novel methodology for the bottom-up manufacture of heuristic composites capable of producing high quality solutions. For this to be done, our methodology unfolds three main steps: execution threads analysis, assemblies characterisation and evolutionary design. In the following, Section~\ref{Sec-SSHH} gives a brief introduction to hyper-heuristics, self-assembly and the context of our investigation. Section~\ref{Sec-PA} enlarges on the proposed approach giving details of the model components and the methodology. After that, experiments and initial results are presented and discussed in Section~\ref{Sec-EAR}. Finally, conclusions and further work are the subject of Section~\ref{Sec-CON}. 

\section{Self-assembly Design and Hyper-heuristics}\label{Sec-SSHH}

Self-assembly is a natural construction process where aggregates emerge spontaneously throughout the progressive interactions among the constituents of the system \cite{gztthesis2008}. Made upon cooperative self-reliant components, self-assembly systems are distributed, not necessarily synchronous, autopoietic mechanisms for the bottom-up fabrication of supra-structures. Polymerisation, nucleic acid structures and crystallisation, to name but a few, are some of the many examples of self-assembly found in nature. 

With the aim to automate the design of a computational model of self-assembly, our previous work has focused on the {\it self-assembly Wang tiles} model which is formally defined as a quintuple $(\mathcal{T}, \Sigma, g, \mathcal{L}, \tau)$ where $\mathcal{T}$ is a finite set of non-empty Wang tiles with square shape, $\Sigma$ is a set of colours which label the edges associated to a tile, $g$ is called the {\it glue function} that evaluates the strength between any two colours of $\Sigma$, $\mathcal{L}$ is a two-dimensional square lattice with interconnected unit square cells and $\tau$ is a positive value that models the kinetic energy of the system \cite{TerGheKenKra2007}. Initially, tiles belonging to $\mathcal{T}$ are randomly located in the lattice, the cells of which can be occupied by one tile at any time. Once tiles are placed on the lattice, they undergo random walks. When two or more tiles collide, the strength between the colours of each pair of colliding edges is evaluated. Subject to this evaluation and the value of the kinetic energy, the associated tiles either self-assemble or remain separated. That is, if the resulting strength is bigger than $\tau$ then the tiles stay still for ever forming an aggregate, otherwise they bounce off. In particular, an evolutionary algorithm, through the process of selection, crossover and mutation, has driven the automated design of Wang tiles capable of self-assembling into a user-defined shape.

Hyper-heuristics are defined as search methodologies that select and combine low-level heuristics to solve hard computational search problems \cite{BurKenNewRosSch2003,Ros2005}. The general aim of a hyper-heuristic is to manufacture unknown heuristics which are fast, well performing and widely applicable to a range of problems. During the process of fabrication, hyper-heuristics receive feedback from the problem domain which indicates how good the chosen heuristics are in relation to the problem at hand and, hence, driving the search process. Although there are some reported conditions under which free lunches are possible \cite{PolGra2009}, hyper-heuristics do not violate the no-free-lunch theorem which indicates that all algorithms that search for optimum of a cost function perform exactly the same when averaged over all possible cost functions, so no algorithm, including hyper-heuristics,  is better than another when considering all optimisation problems. Studying novel approaches for the development of hyper-heuristics is important since they are domain-independent problem strategies that operate on a space of heuristics, rather than on a space of solutions, and rise the level of generality on automated problem solving. Hyper-heuristics have been employed for solving search and optimisation problems such as bin-packing \cite{RosSchMarHar2002,BurHydKen2006}, timetabling \cite{PilBan2008}, scheduling \cite{CowKenHan2002,CowCha2003} and satisfiability \cite{BadPol2007} among others. For detailed reviews of hyper-heuristics and their applications, please refer to \cite{Ros2005,OzcBilKor2008,ChakCow2009,BurKenNewRosSch2003}.

The automated manufacture of heuristic search strategies by means of hyper-heuristics has received increasing attention in the last ten years or so. Recent investigations have sprung up in two main different flavours of hyper-heuristics: 1) heuristics that choose heuristics and 2) heuristics that generate heuristics. In the first case, a learning mechanism assists the selection of low-level heuristics according to their historical performance during the search process. In the second case, the focus is on the search of components that once combined generate a new heuristic suitable for the problem at hand. Approaches based on genetic programming have been proposed for the automated generation of heuristics \cite{BurHydKenWoo2007,OltDum2004,CowKenHan2002}. From an engineering point of view, we believe that the manufacture of aggregates resulting from local interactions among autonomous cooperative entities is an interesting route for developing a new alternative within the second flavour of hyper-heuristics. That is, our interest is on applying self-assembly as a mechanism to develop a hyper-heuristic approach to then automatically generate problem-specific good performing heuristics. An early idea of program constructions by means of self-assembly is reported in \cite{LinKraGar2006} where the automated self-assembly programming paradigm (ASAP$^2$) is introduced as a self-assembly model for unconventional computing. ASAP$^2$ is defined in terms of gas' molecules embodying portions of software sampled from man-made program libraries. Thus, software molecules interact to one another subject to different values of temperature and pressure giving rise to a variety of program architectures. The purpose of the present investigation is to apply previous experiences in evolutionary design of self-assembly Wang tiles in order to develop a nature-inspired cooperative strategy for the automatic construction of heuristic search strategies.

\section{Proposed Approach}\label{Sec-PA}

This section presents our approach for the automated design of heuristics by means of self-assembly. In the first part, we introduce the motivation, the components of the model and the general aim of our strategy. In the second part, we give a detailed step-by-step description of the methodology together with the goals associated to each stage.

\subsection{Model}

\begin{figure}[ht]
\centering
\begin{tabular}{c@{\extracolsep{50pt}}c}
\includegraphics[scale=0.30]{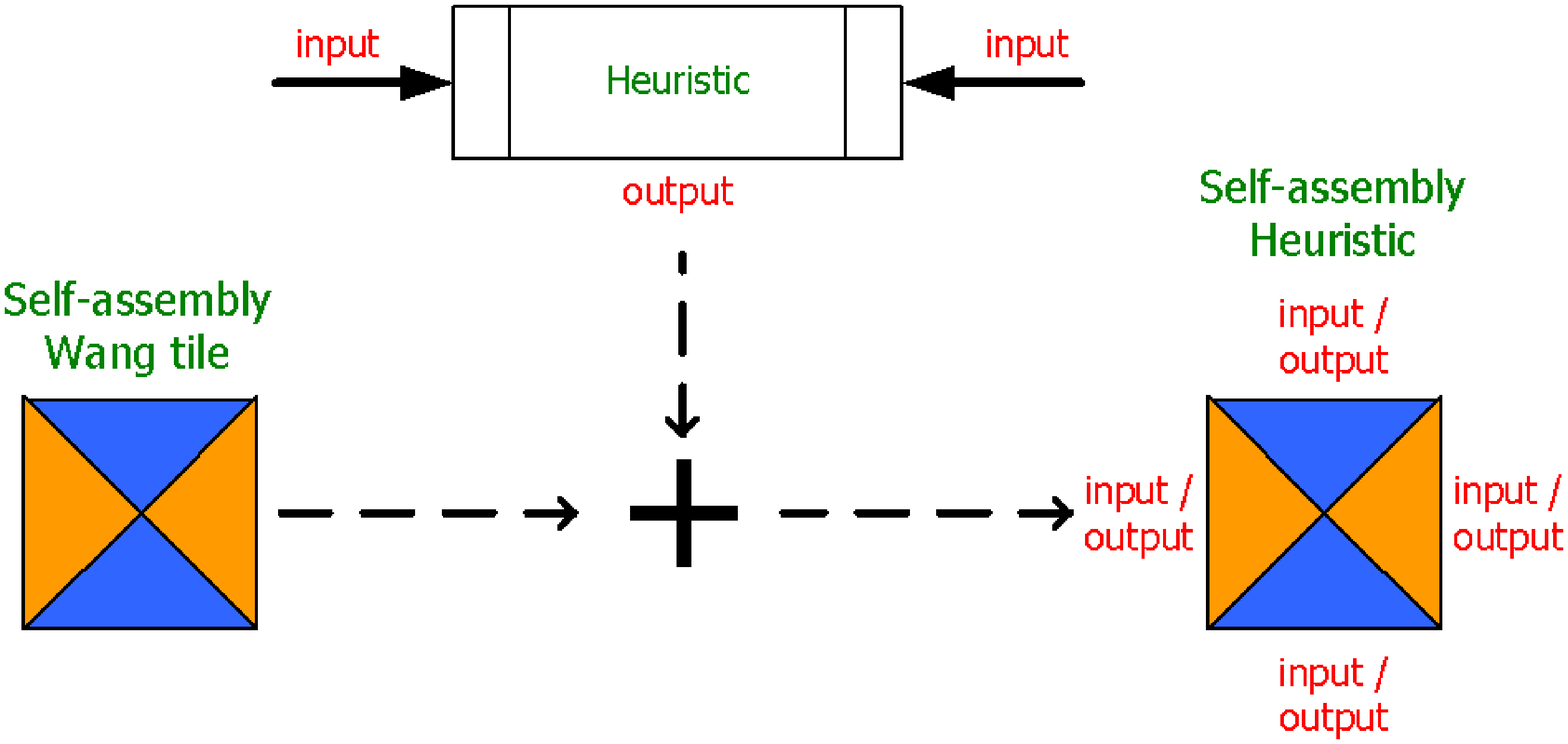} & \includegraphics[scale=0.30]{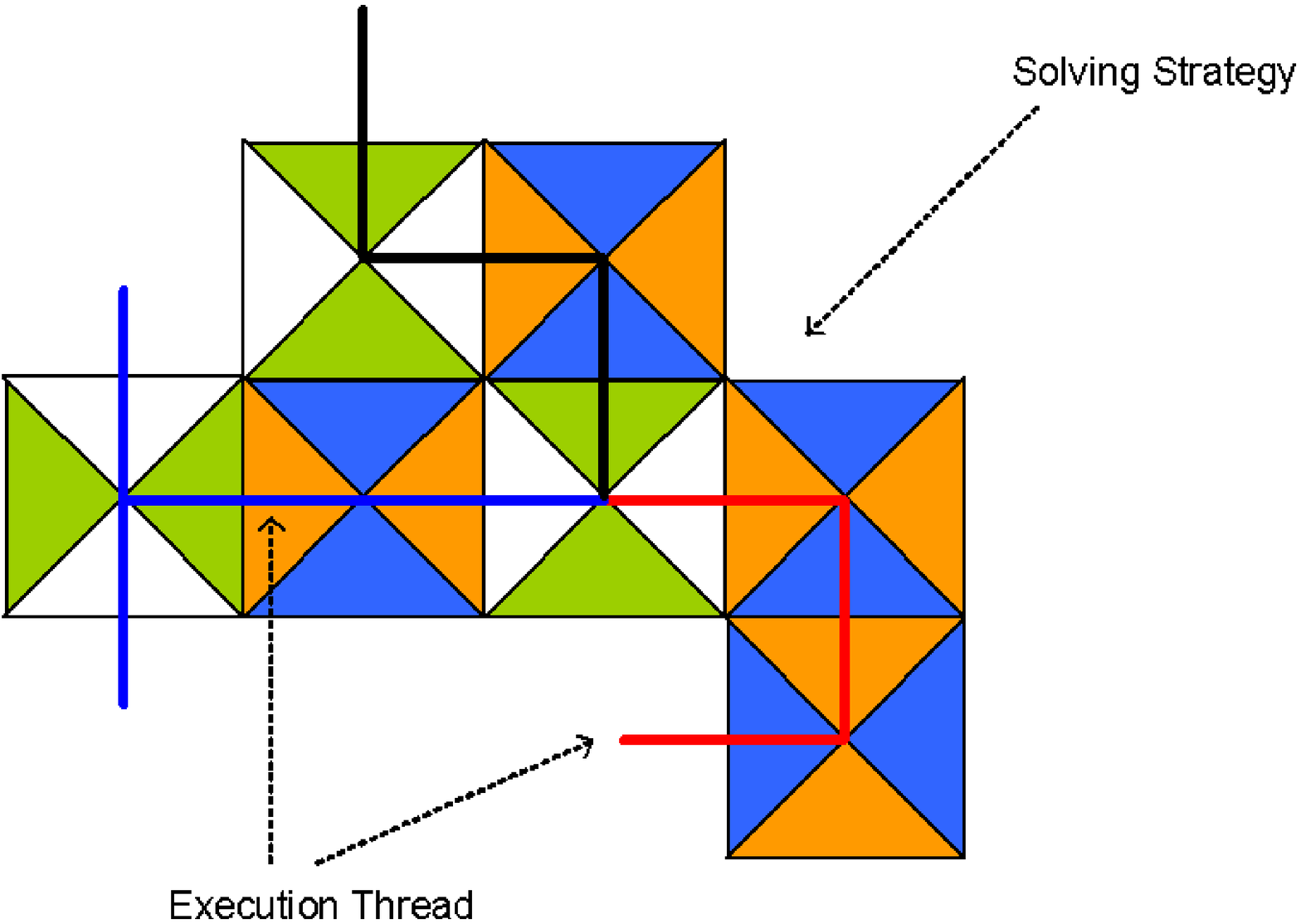}\\
(a) & (b)
\end{tabular}
\caption{\label{sa_wang_tiles_embeding_heuristics}A self-assembly Wang tile embedding a heuristic with two inputs and an output (a); an aggregate defining a composition of self-assembly heuristics with two alternative execution threads comprising five heuristics each (b).}
\end{figure}

Inspired by the physical process of crystallization, Winfree has introduced the Tiles Assembly Model \cite{winfree98simulations} as a quadruple $(\mathcal{T}, t_s, g, \tau)$ where $\mathcal{T}$ is a finite set of non-empty tile types, $t_s$ is a seed tile belonging to $\mathcal{T}$, $g$ is a strength function and $\tau$ is a threshold parameter.  This model has proven to have computational power by simulating a one-dimensional blocked cellular automata. This simulation demonstrates that a unique pattern is always produced, regardless of the order in which tiles are aggregated, and that such arrangement represents information ultimately modified by tile additions interpreted as rewrite rules. Winfree's model has been employed for solving NP-hard problems \cite{Brun08a,Brun08b,1355014} proving that tile structures interpreted as programs are in fact successful. The aim of our model lies in the automatic development of problem solving entities. To be more precise, we consider a self-assembly Wang tile as an independent {\it low-level heuristic} (Figure \ref{sa_wang_tiles_embeding_heuristics}(a)) and an aggregate as a {\it solving strategy} in full (Figure \ref{sa_wang_tiles_embeding_heuristics}(b)).  
\\

Although there can be sophisticated ways to decide which is the input/output of a tile and how to execute the low-level heuristics of an aggregate, we prefer to adopt a simple alternative based on the construction of sequences of low-level heuristics. We refer to these sequences as {\it execution threads} (Figure \ref{sa_wang_tiles_embeding_heuristics}(b)), the construction process of which and the way they operate are given in Section \ref{ETC}. Thus, given an instance of a combinatorial optimisation problem and a set of low-level heuristics embedded in self-assembly Wang tiles, the questions pursued by our research are:\\

{\it Is it possible to automatically design an assembly of heuristics, the execution threads of which help to find optimal solutions for a given combinatorial optimisation problem?}\\

{\it If the answer to the above question is yes, is it possible to reuse the methodology in order to tackle a different combinatorial optimisation problem?}\\

In order to address these questions, we propose a methodology divided in three main stages: 1) execution thread analysis, 2) assembled heuristic characterisation and 3) evolutionary design. These stages are described in more detail in the following subsection.

\subsection{Methodology}\label{subsec-MTH}

The purpose of the \emph{execution thread analysis} in stage 1 is to shed light on common combinations of heuristics that help to produce high quality solutions when applied to a given problem instance. Hence, given a set of execution threads, the research question to address in stage 1 is:\\

{\it Is it possible to identify common combinations of heuristics? If yes, how do they look like and how reliable is the performance of such combinations when applied to different instances of the problem at hand?}\\

After the analysis and assessment of the execution threads, the best ranked ones are selected and their associated assembled heuristics are used as input to stage 2. Then, the goal of the \emph{assembled heuristics characterisation} in stage 2 is to define the target shape that the self-assembly system should attempt to generate. Thus, given a set of assembled heuristics the question to address in stage 2 is:\\

{\it Which is the morphology characterizing the high quality assembled heuristics?}\\

The findings reported in \cite{TerGheKenKra2007,gztthesis2008} recognise the application of evolutionary algorithms as suitable mechanisms for the automated design of Wang tiles capable of self-assembling in a user defined target shape. Therefore, given a target shape and a set of low-level heuristics embedded in a self-assembly Wang tiles system, the aim of the \emph{evolutionary design} in stage 3 is to address the following question:\\

{\it Is it possible to conduct an automated design of a set of low-level heuristics, the assemblages of which return high quality solutions when applied to a given combinatorial optimisation problem instance?}\\

The above methodology is expected to produce a novel procedure for the automated construction of tailored effective and efficient heuristic search strategies. This would also bring additional evidence to support the claim that cooperative strategies found in nature are robust mechanisms suitable for the development of solutions to combinatorial optimisation problems.

\section{Experiments and Results}\label{Sec-EAR}

This section presents the preliminary findings obtained by stage 1 of the above methodology. The chosen combinatorial optimisation problem is the widely known Travelling Salesman Problem (TSP) in its symmetric version. As this paper presents a proof of concept, a relatively easy to solve instance is employed to illustrate the concept of self-assembly hyper-heuristics. The TSP instance considered here is kroA100 which comprises $100$ cities distributed in the Euclidean space. The objective value corresponding to the known optimum solution (shortest tour) for this instance is $21282$ (see TSPLIB\footnote{http://elib.zib.de/pub/mp-testdata/tsp/tsplib/tsplib.html}). For the experiments in this paper, we take the known optimum tour of kroA100 and apply $200$ random swaps in order to generate a `disturbed' tour. In this way, we generated $10$ different `disturbed' tours which are then used to evaluate the performance of the execution threads (combinations of heuristics).
\begin{algorithm}[H]
  \caption{ExecutionThreadsAnalysis}
  \label{pseudocode}
  \algsetup{
	indent=2em,
	linenodelimiter=.
  }
  \begin{algorithmic}[1]
    \REQUIRE $PI$ a symmetric TSP instance
    \STATE $ETS_c \leftarrow$ collect $N$ execution threads
    \FORALL{execution threads $ET_i$ in $ETS_c$}
		\STATE apply$(ET_i, PI, times)$
	\ENDFOR
    \STATE $ETS_f \leftarrow$ filter best $ETS_c$
    \STATE $PATTERNS \leftarrow$ analyse common heuristics $ETS_f$
    \STATE $csET \leftarrow$ build a $PATTERNS$-based execution thread 
    \FORALL{execution threads $ET_i$ in $ETS_f \cup \{csET\}$}
		\STATE $ETS_r \leftarrow$ generate $300$ random execution threads
		\FORALL{execution threads $ET_j$ in $ETS_r$}
			\STATE apply$(ET_i, PI, times)$
			\STATE apply$(ET_j, PI, times)$
		\ENDFOR
	\ENDFOR
	\STATE assess $ETS_f$ vs. $ETS_r$
  \end{algorithmic}
\end{algorithm}
	
The \emph{execution threads analysis} stage involves 3 steps: execution threads collection, detection of patterns of heuristics and performance evaluation. For this, we developed the procedure shown in Algorithm \ref{pseudocode} where lines $1$ to $5$ define the first step, lines $6$ to $7$ specify the second step and the remaining lines outline the performance evaluation step. Each of these steps is described in more detail in the following subsections.

\subsection{Execution Threads Collection}\label{ETC}

\begin{figure}[ht]
\centering
\begin{tabular}{c@{\extracolsep{50pt}}c}
\includegraphics[scale=0.32]{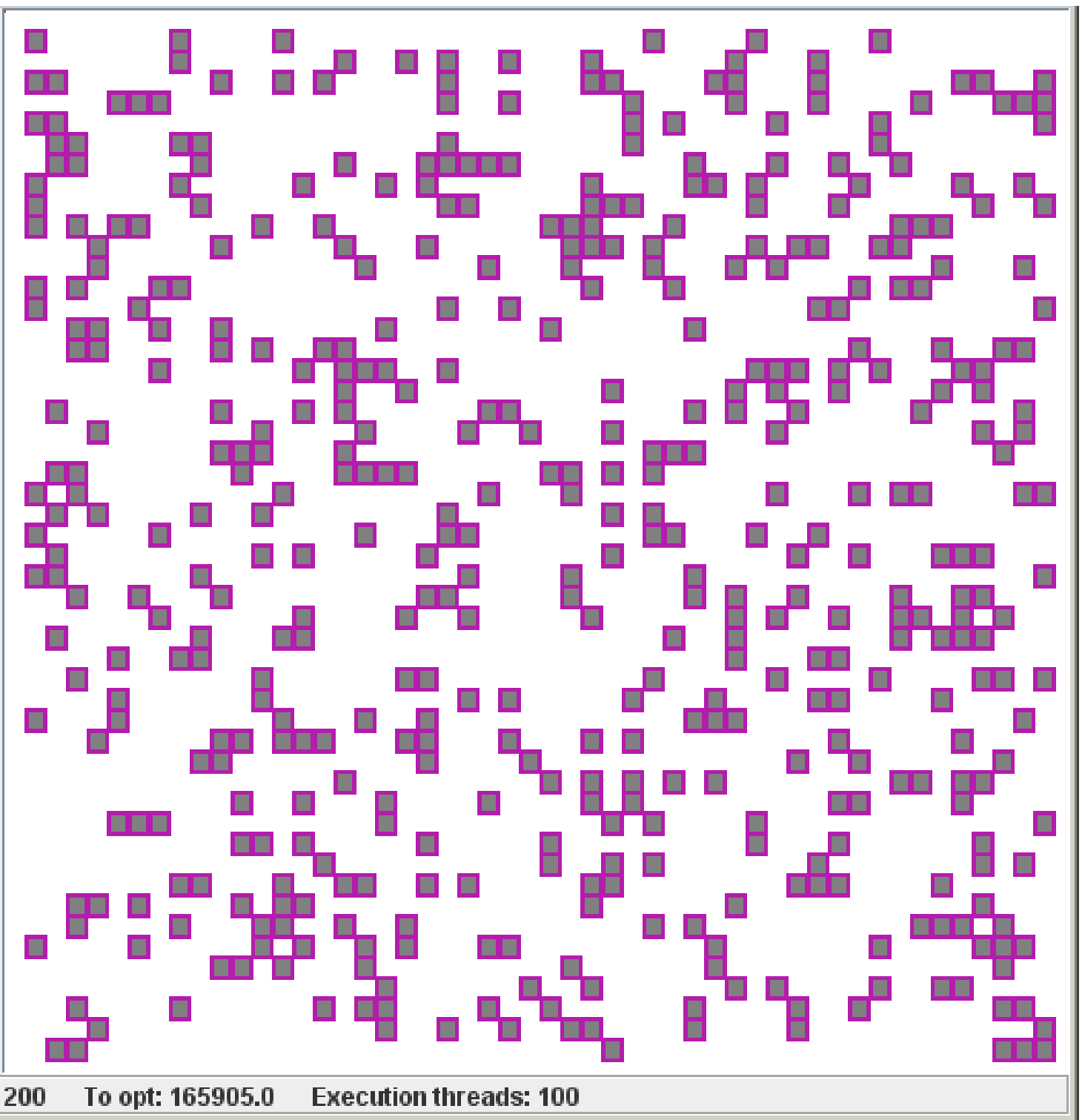} & \includegraphics[scale=0.32]{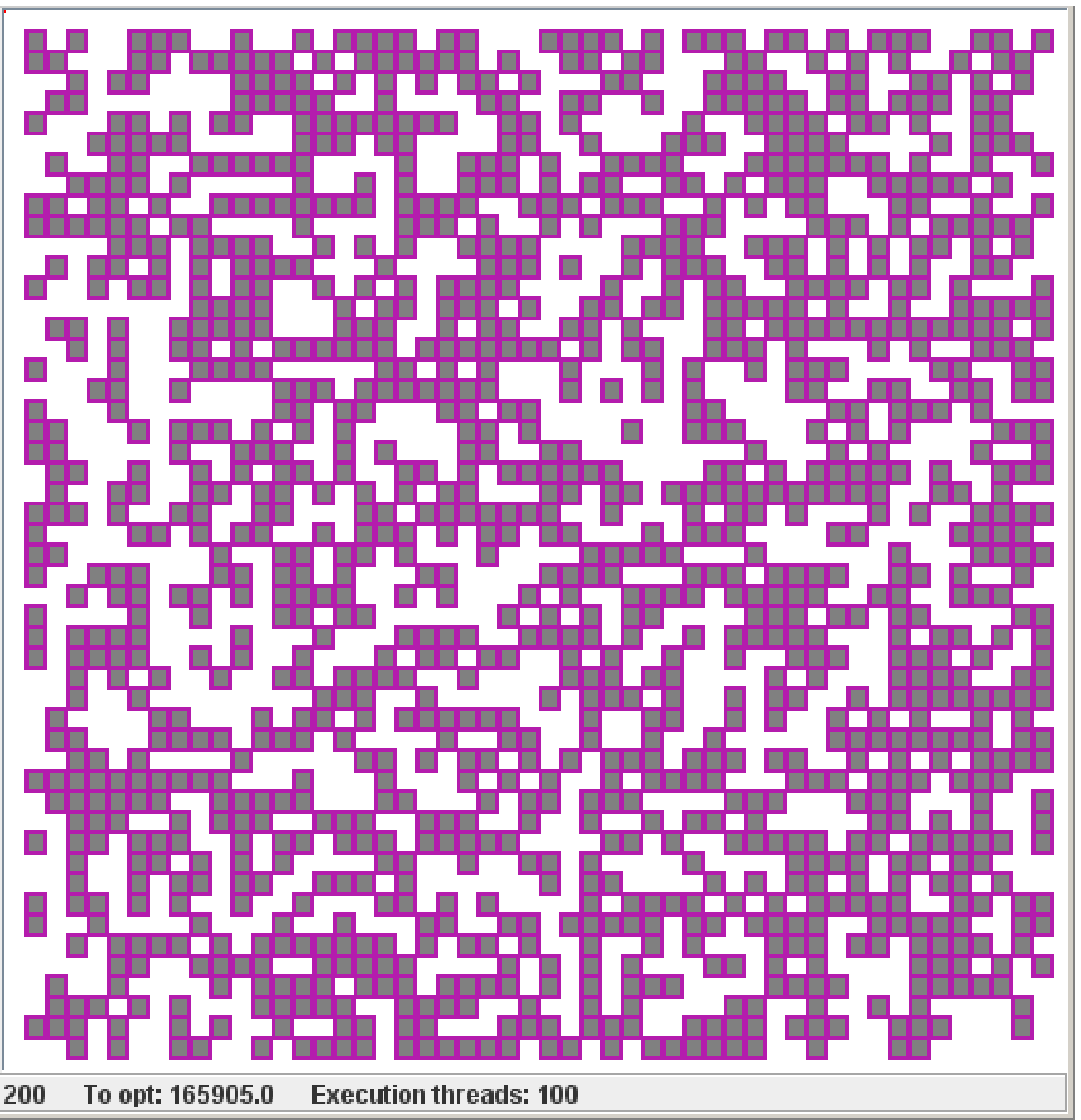}\\
(a) & (b)
\end{tabular}
\caption{\label{percolation_clusters}Two percolation clusters on a two-dimensional square lattice using $\rho = 0.2$ (a) and $\rho = 0.5$ (b). The low-level heuristics are uniformly distributed across the percolation cluster.}
\end{figure}

In order to generate the execution threads, we employed a percolation cluster model. A percolation model is defined as a collection of both empty and occupied sites distributed across a lattice. In this model, the non-empty sites of the lattice can be partitioned into clusters such that there is always a path between any two sites of the same cluster and non-empty sites of different clusters are disconnected. In particular, we employed a 2D square site percolation cluster with occupation probability $\rho \in \{0.05, 0.1, 0.15, \ldots, 0.95\}$. That is, for each site of the cluster a random value $v \in [0, 1]$ is obtained. If $v \leq  \rho$ then the site is filled with a low-level heuristic chosen at random, otherwise it stays empty. Figure \ref{percolation_clusters} shows two percolation clusters generated with $\rho = 0.2$ and $\rho = 0.5$. 

\begin{figure}[ht]
\centering
\includegraphics[scale=0.30]{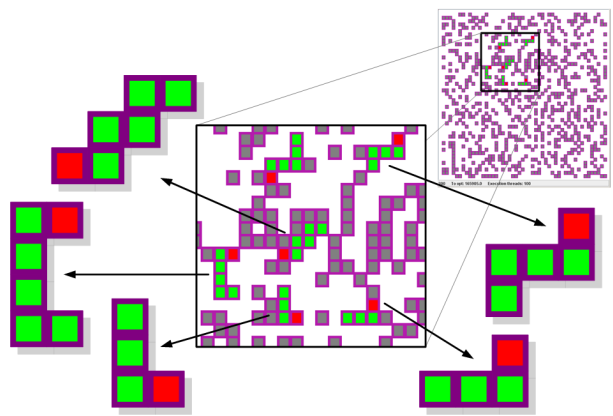}
\caption{\label{execution_thread}Sample execution threads collected from a percolation cluster. Red sites indicate the initial heuristic from where the execution threads start. Green sites are the successive heuristics collected as the random walk is performed.}
\end{figure}

The reason for using percolation clusters is that the geometry (shape) of the resulting aggregates is similar to the ones resulting from the self-assembly Wang tiles system. In addition, since there is a correlation between the occupation probability $\rho$ (size of the aggregates) and the length of the possible random walks \cite{PaiJolNeoWWW}, different values of $\rho$ would allow us to systematically explore different lengths of execution threads. Therefore, in order to collect an execution thread, a random walk over the lattice is performed. That is, a non-empty site in the lattice is arbitrary selected and from there the nearest one is chosen at random. If the selected site hosts a heuristic, the execution thread increments its length and the process is repeated by choosing the following nearest site. Otherwise, the execution thread does not increment its length and the collection finishes. Note that backward walks and crossings could increment the length of the execution thread by adding occurrences of the already collected heuristics. Figure \ref{execution_thread} shows an example of five execution threads collected from a site percolation cluster with $\rho = 0.5$.\\

For the chosen problem, the low-level heuristics used here are local searches for the TSP that can be deterministic (e.g. always selecting the best of a set of improving two-edges interchange) or stochastic (e.g. selecting at random from a set of improving two-edges interchange, and hence potentially giving different results if re-executed). In particular, {\it 2-Opt}, {\it 3-Opt}, {\it OR-Opt} and {\it Node Insertion} are deterministic whereas {\it 1-city Insertion}, {\it 2-exchange}, {\it Arbitrary Insertion} and {\it Inver-over} are stochastic. These eight low-level heuristics are originally defined in \cite{Babin2007,Brest2005,KraSmi2007,Reinelt1994,Tao1998} and can be summarised as follows:\\
\\
{\bf 2-Opt}, that eliminates and reconnects those two edges which best minimise the length of the tour.\\
\\
{\bf 3-Opt}, that eliminates and reconnects those three edges which best minimise the length of the tour.\\
\\
{\bf OR-Opt}, that eliminates and reinserts a sub-tour of three consecutive cities to the best location, then eliminates and reinserts a sub-tour of two consecutive cities to the best location and, finally, eliminates and reinserts a sub-tour of one city to the best location.\\
\\
{\bf Node Insertion}, that removes and reinserts the city which best minimises the length of the tour.\\
\\
{\bf $k$-city Insertion}, that removes a sub-tour $s$ beginning with a randomly chosen city $i$ through city $i+k$, connects $i-1$ to $i+k+1$ and reinserts $s$ elsewhere in the tour. In particular, we set $k = 1$.\\
\\
{\bf $n$-exchange}, that removes $n$ edges and reinserts $n$ new edges to rebuild a feasible tour. In particular, we set $n = 2$.\\
\\
{\bf Arbitrary Insertion}, that removes a sub-tour $S$ beginning with a randomly chosen city $i$ through a randomly chosen city $j$, connects city $i-1$ with city $j+1$ and reinserts each city of $S$ in the cheapest possible way.\\
\\
{\bf Inver-over}, that removes a sub-tour $S$ beginning with a randomly chosen city $i$ through a randomly chosen city $j$, and reinserts $S$ inverted, i.e. connecting city $i-1$ to city $j$ and city $i$ to city $j+1$.\\

In our case, the low-level heuristics described above operate in a hill climber style \cite{OzcBilKor2006}, i.e. there is always an improvement on the TSP where they apply.

\subsection{Detection of Patterns of Heuristics}

\begin{figure}[ht]
\centering
\includegraphics[scale=0.40]{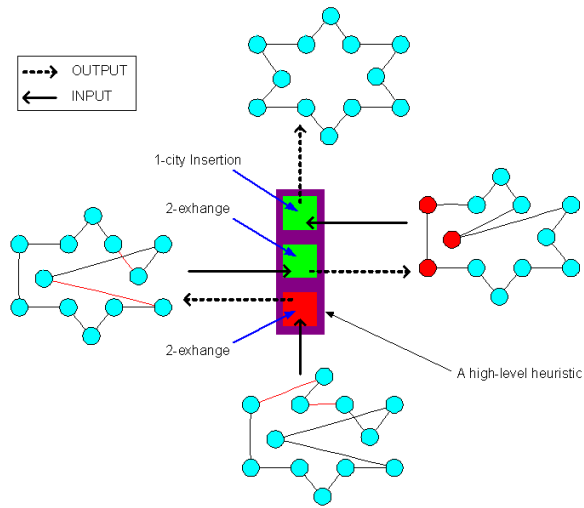}
\caption{\label{application_of_heuristics}An execution thread in which applications of two 2-exchange heuristics and a 1-city insertion heuristic find the optimum solution for the Star of David tour. Edges and cities where the three low-level heuristics apply are coloured in red.}
\end{figure}

For each experiment, we generated $50$ percolation clusters with each of the nineteen values of $\rho$. From each of the resulting clusters, $100$ execution threads were collected and applied to the same problem instance. In this context, an execution thread application is seen as a pipeline process in which the chain of processing elements is given by the sequence of low-level heuristics and the information to be processed is the problem instance. Thus, heuristics are applied one after another in the order in which they appear in the execution thread and producing better or equal solutions at each step. With the purpose to illustrate this process, Figure \ref{application_of_heuristics} depicts a very simple example in which an execution thread, comprising of 1-city insertion and 2-exchange heuristics, is applied to a TSP instance. Red tile indicates the chosen initial heuristic, where the execution thread starts, and the green tiles the successive collected ones. Therefore, the application of the execution thread begins by applying the heuristic embedded in the red tile and follows by applying the heuristics embedded in the rest of the consecutive green tiles.\\

\begin{figure}[ht]
\centering
\includegraphics[scale=0.40]{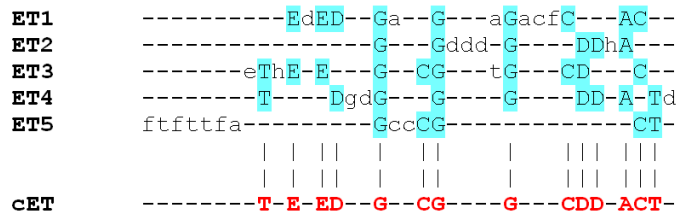}
\caption{\label{results_muscle}Multiple sequence alignment of the top five execution threads collected from percolation clusters generated with $\rho = 0.9$. Capitals highlighted in blue indicate the common sequences of heuristics.}
\end{figure}

In order to identify common combinations of heuristics within the collected execution threads, we grouped the execution threads according to the $\rho$ that generated their underlying percolation cluster. Within each group, the execution threads are then sorted according to the distance between the solution that the execution thread produces and the known optimum solution. The top five execution threads within each group are then selected and encoded as sequences of characters using `A' to represent 1-city insertion, `C' to represent 2-opt, `D' to represent 3-opt, `E' to represent OR-opt, `T' to represent 2-exchange, `F' to represent node insertion, `G' to represent arbitrary insertion and `H' to represent inver-over. Hence, in order to identify common combinations of heuristics among the filtered execution threads, we employ a multiple sequence alignment (MSA) method \cite{SetMei1997} over the encodings. The results reveal that there are indeed occurrences of common combinations, i.e. patterns of heuristics, among the best ranked execution threads. Thus, these findings give a positive answer to the first research question that we stated for stage 1 of our methodology (Section~\ref{subsec-MTH}). Figure \ref{results_muscle} highlights in blue the patterns found among the best five execution threads collected from percolation clusters using $\rho = 0.9$.

An extra execution thread is then constructed in terms of the spotted patterns of heuristics. This procedure consists in copying the matching characters between two or more encodings into a new sequence from left to right and following the position in which they appear. For instance, Figure \ref{results_muscle} shows that cET is the resulting pattern-based execution thread encoded as TEEDGCGGCDDACT, after combining the common patterns from the input execution threads ET1 to ET5. Given that this execution thread is built in terms of common combinations of heuristics, its performance is then expected to be as good as (or better than) any of the top ranked. Notice that the length of the constructed execution thread varies according to the number of matches. Since this is related to the way in which the construction procedure is defined, we left open to further investigations other alternatives to construct the common execution thread, e.g. by calculating the optimal common sequence in the alignment.

\subsection{Performance Evaluation}

Since the best five execution threads were evaluated only once, a better way to assess their performance is needed. For that reason, we assess the best five execution threads, and the one constructed in terms of patterns of heuristics, by conducting a vis-a-vis comparison between their performances and randomly generated ones with the hope that, on average, the best tour improvements are obtained by the common-sequence execution threads. Thus, for each of the six execution threads, $300$ copies are obtained and for each of these copies a new execution thread equal in length is randomly created. Since stochastic low-level heuristics could be part of an execution thread, a total of $10$ independent evaluations are performed and the average distance between the lengths of the resulting tours and the known optimum was considered as the measure of its performance. 
\begin{figure}[ht]
\centering
\includegraphics[scale=0.51, angle=-90]{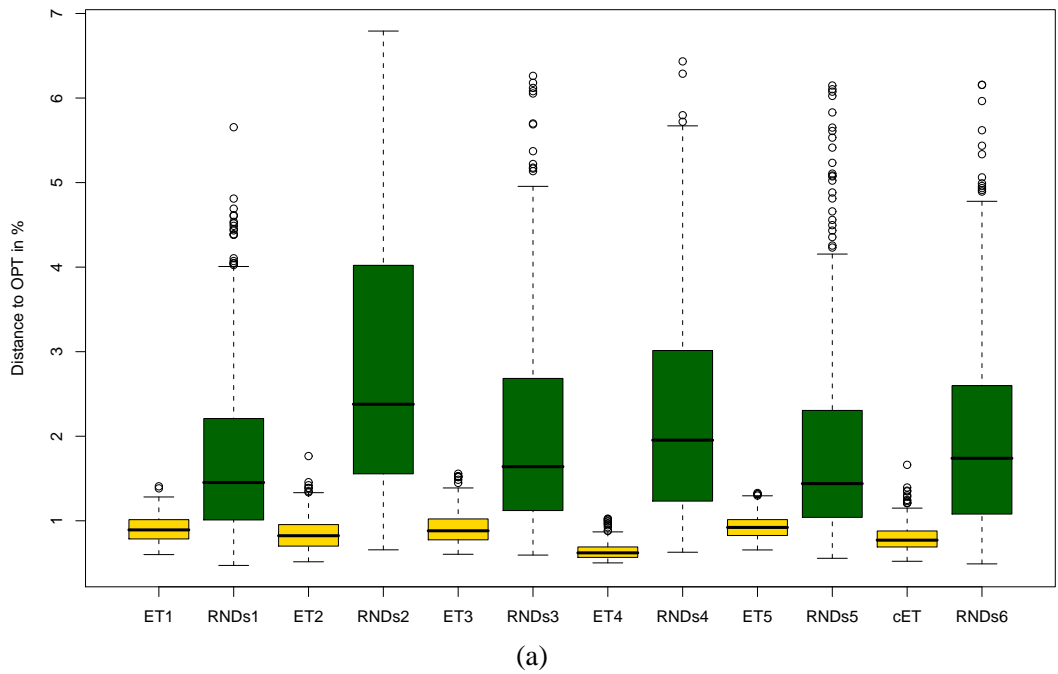}\\
(a)\\
\includegraphics[scale=0.51, angle=-90]{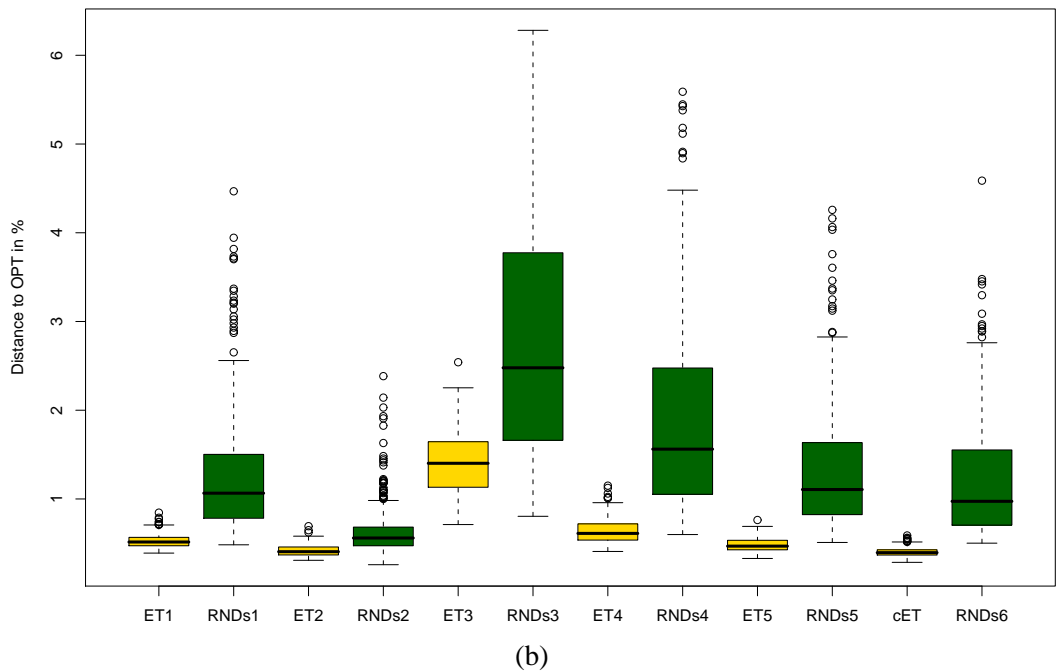}\\
(b)
\caption{\label{both_trends}Performance evaluation of two independent experiments. Each boxplot summarises a vis-a-vis comparison between the performances of the best ranked execution threads (ET1, ET2, ET3, ET4, ET5) and their associated randomly generated (RNDs1, RNDs2, RNDs3, RNDs4, RNDs5) as well as the pattern-based execution thread (cET) and its associated randomly generated (RNDs6).}
\end{figure}

Two representative analyses of the performance evaluation results obtained across the $10$ experiments are shown in Figure \ref{both_trends} (a - b). The boxplots in Figure \ref{both_trends} (a) depict the assessment when employing a TSP tour with value $165905$ whilst the boxplots in Figure \ref{both_trends} (b) correspond to the results for a TSP tour with value $191550$. In general, we can observe that in both experiments the best five execution threads (ET1, ET2, ET3, ET4, ET5) outperform on average the associated randomly generated ones (RNDs1, RNDs2, RNDs3, RNDs4, RNDs5). In particular, some of the smallest observations depicted by the boxplots of RNDs1 and RNDs5 in Figure \ref{both_trends} (a) show that few randomly generated execution threads outperformed ET1 and ET5 in Figure \ref{both_trends} (a). A similar situation can be observed between RNDs2 in Figure \ref{both_trends} (b)  and ET2 in Figure \ref{both_trends} (b). This is not surprising since among all the available heuristics, there is still the chance that certain arrangements of such were not considered during the execution threads collection step. Nevertheless, according to the values of the medians, the amount of missing arrangements found among the $300$ randomly generated is still not enough to outperform in average a systematically collected execution thread.

Regarding the execution threads constructed in terms of patterns of heuristics (cET), it is clear that their performance is better when compared to the associated randomly generated execution threads (RNDs6). In addition, it is always the case that the performance of an cET (common-sequence execution thread) is as competitive as the performance of the best five input execution threads. Hence, these findings indicate that the discovered patterns are in fact beneficial combinations of low-level heuristics necessary for solving a symmetric TSP instance. All in all, the analyses and results unfolded here constitute a positive answer to the second research question stated for the first stage of the presented methodology (Section~\ref{subsec-MTH}), i.e. the identified common-sequences of heuristics are indeed reliable.

\clearpage

\section{Conclusions}\label{Sec-CON}

In this paper, we proposed a nature-inspired approach for the automated design of heuristics following the rationale of hyper-heuristics which are heuristic methods to generate tailored heuristics for the problem in hand. Our model considers the use of self-assembly Wang tiles embedding low-level heuristics and their assemblages as higher-level heuristic strategies. The proposed methodology consists of 3 stages: execution thread analysis, assembled heuristics characterisation and evolutionary design.

In particular, we reported experiments and results from the \emph{execution threads analysis} stage involving three steps: execution threads collection, detection of patterns of heuristics and performance evaluation. On the one hand, the initial findings confirm that there are indeed common patterns of heuristics among the top ranked execution threads. This emergent recurrent structures are non-divisible local search strategies beneficial to achieve good solutions when solving a symmetric TSP instance. On the other hand, the assessment of the execution threads produced positive results about the reliability, with respect to the performance, of the collected local search strategies. These findings reveal that the top execution threads are good performing arrangements of heuristics and that the emergent patterns are beneficial to obtain good solutions. 

To continue with our methodology, future work involves the morphological characterisation of the common-sequence assembled heuristics and the evolutionary design. The integration of these two stages together with the methodology presented here is expected to produce a novel procedure for the automated construction of heuristic search strategies.

\section{Acknowledgements} 
The research reported in this work is funded by EPSRC grant (EP/D061571/1) {\it Next Generation Decision Support: Automating the Heuristic Design Process}. 

\bibliographystyle{eptcs}

\begin{thebibliography}{1}

\bibitem{Babin2007}
G.~Babin, S.~Deneault, and G.~Laporte.
\newblock Improvements to the or-opt heuristic for the symmetric traveling
  salesman problem.
\newblock {\em Journal of the Operational Research Society}, (58):402--407,
  2007.

\bibitem{BadPol2007}
M.~B. Bader-El-Den and R.~Poli.
\newblock A gp-based hyper-heuristic framework for evolving 3-sat heuristics.
\newblock In {\em Genetic and Evolutionary Computation Conference}, pages
  1749--1749. ACM, 2007.

\bibitem{Brest2005}
J.~Brest and J.~Zerovnik.
\newblock A heuristic for the asymmetric traveling salesman problem.
\newblock In {\em Metaheuristics International Conference}, pages 145--150,
  2005.

\bibitem{Brun08a}
Y.~Brun.
\newblock Constant-size tileset for solving an {NP}-complete problem in
  nondeterministic linear time.
\newblock In {\em {DNA} Computing}, volume 4848, pages 26--35. Springer Berlin
  / Heidelberg, 2008.

\bibitem{Brun08b}
Y.~Brun.
\newblock Reducing tileset size: 3-{SAT} and beyond.
\newblock In {\em {DNA} Computing}, page 178, 2008.

\bibitem{1355014}
Y.~Brun.
\newblock Solving np-complete problems in the tile assembly model.
\newblock {\em Theor. Comput. Sci.}, 395(1):31--46, 2008.

\bibitem{BurKenNewRosSch2003}
E.~K. Burke, E.~Hart, G.~N. Kendall, J.~Newall, P.~Ross, and S.~Schulenburg.
\newblock {\em Handbook of Meta-Heuristics}, chapter Hyper-Heuristics: An
  Emerging Direction in Modern Search Technology, pages 457--474.
\newblock Kluwer, 2003.

\bibitem{BurHydKen2006}
E.~K. Burke, M.~R. Hyde, and G.~Kendall.
\newblock Evolving bin packing heuristics with genetic.
\newblock In {\em Parallel Problem Solving from Nature}, volume 4193, pages 860--869. Springer-Verlag, 2006.

\bibitem{BurHydKenWoo2007}
E.~K. Burke, M.~R. Hyde, G.~Kendall, and J.~Woodward.
\newblock Automatic heuristic generation with genetic programming: evolving a
  jack-of-all-trades or a master of one.
\newblock In {\em Genetic and Evolutionary Computation Conference}, pages
  1559--1565. ACM, 2007.

\bibitem{ChakCow2009}
K.~Chakhlevitch and P.~I. Cowling.
\newblock Hyperheuristics: Recent developments.
\newblock In {\em Adaptive and Multilevel Metaheuristics}, volume 136, pages
  3--29. Springer, 2008.

\bibitem{CowCha2003}
P.~Cowling and K.~Chakhlevitch.
\newblock Hyperheuristics for managing a large collection of low level
  heuristics to schedule personnel.
\newblock In {\em IEEE Congress on Evolutionary Computation}, pages 1214--1221.
  IEEE Computer Society, 2003.

\bibitem{CowKenHan2002}
P.~Cowling, G.~Kendall, and L.~Han.
\newblock An investigation of a hyperheuristic genetic algorithm applied to a
  trainer scheduling problem.
\newblock In {\em IEEE Congress on Evolutionary Computation}, pages 1185--1190.
  IEEE Computer Society, 2002.

\bibitem{PaiJolNeoWWW}
S.~J. G.~M.~Paily and S.~Neogi.
\newblock Two dimensional random walk on percolation clusters.
\newblock Available at
  \url{http://www.personal.psu.edu/saj169/PercolationRW/PercolationRw.html}.

\bibitem{KraSmi2007}
N.~Krasnogor and J.~Smith.
\newblock Memetic algorithms: The polynomial local search complexity theory
  perspective.
\newblock {\em Journal of Mathematical Modelling and Algorithms}, 7:3--24,
  2008.

\bibitem{LinKraGar2006}
L.~Li, J.~Garibaldi, and N.Krasnogor.
\newblock Automated self-assembly programming paradigm: initial investigation.
\newblock In {\em IEEE International Workshop on Engineering of Autonomic
  and Autonomous Systems}, pages 25--36. IEEE, 2006.

\bibitem{OltDum2004}
M.~Oltean and D.~Dumitrescu.
\newblock Evolving tsp heuristics using multi expression programming.
\newblock In {\em Conference on Computational Science},
  volume 3037, pages 670--673, 2004.

\bibitem{OzcBilKor2006}
E.~\"{O}zcan, B.~Bilgin, and E.~Korkmaz.
\newblock Hill climbers and mutational heuristics in hyperheuristics.
\newblock In {\em Parallel Problem Solving from  Nature}, pages 202--211, 2006.

\bibitem{OzcBilKor2008}
E.~\"{O}zcan, B.~Bilgin, and E.~E. Korkmaz.
\newblock A comprehensive analysis of hyper-heuristics.
\newblock {\em Intell. Data Anal.}, 12(1):3--23, 2008.

\bibitem{PilBan2008}
N.~Pillay and W.~Banzhaf.
\newblock A study of heuristic combinations for hyper-heuristic systems for the
  uncapacitated examination timetabling problem.
\newblock {\em European Journal of Operational Research}, 197(2):482--491,
  2009.

\bibitem{PolGra2009}
R.~Poli and M.~Graff.
\newblock There is a free lunch for hyper-heuristics, genetic programming and
  computer scientists.
\newblock In {\em European Conference on Genetic Programming}, pages
  195--207. Springer-Verlag, 2009.

\bibitem{Reinelt1994}
G.~Reinelt.
\newblock {\em The traveling salesman: Computational solutions for {TSP}
  applications}.
\newblock Springer-{V}erlag, 1994.

\bibitem{Ros2005}
P.~Ross.
\newblock {\em Hyper-heuristics}, pages 529--556.
\newblock Springer, 2005.

\bibitem{RosSchMarHar2002}
P.~Ross, S.~Schulenburg, J.~G. Mar\'{\i}n-Bl\'{a}zquez, and E.~Hart.
\newblock Hyper-heuristics: Learning to combine simple heuristics in
  bin-packing problems.
\newblock In {\em Genetic and Evolutionary Computation Conference}, pages
  942--948. Morgan Kaufmann Publishers Inc., 2002.

\bibitem{SetMei1997}
J.~Setubal and J.~Meidanis.
\newblock {\em Introduction to Computational Molecular Biology}.
\newblock PWS Publishing, 1997.

\bibitem{Tao1998}
G.~Tao and Z.~Michalewicz.
\newblock Inver-over operator for the tsp.
\newblock In {\em Parallel Problem Solving from  Nature}, pages 803--812. Springer-Verlag, 1998.

\bibitem{gztthesis2008}
G.~Terrazas.
\newblock {\em Automated Evolutionary Design of Self-Assembly and
  Self-Organising Systems}.
\newblock PhD thesis, University of Nottingham, 2008.

\bibitem{TerGheKenKra2007}
G.~Terrazas, M.~Gheorghe, G.~Kendall, and N.~Krasnogor.
\newblock Evolving tiles for automated self-assembly design.
\newblock In {\em IEEE Congress on Evolutionary Computation}, pages 2001--2008.
  IEEE Computer Society, 2007.

\bibitem{winfree98simulations}
E.~Winfree.
\newblock Simulations of computing by self-assembly.
\newblock In {\em DNA-Based Computers}, pages 213--242,
  1998.

\end{thebibliography}

\end{document}